\documentclass[conference]{IEEEtran}
\IEEEoverridecommandlockouts
\usepackage{amsmath,amssymb,amsfonts}
\usepackage{algorithmic}
\usepackage{graphicx}
\usepackage{textcomp}
\usepackage{xcolor}
\usepackage[english]{babel}
\usepackage{times}
\usepackage{epsfig}
\usepackage{graphicx}
\usepackage{cite, url}
\usepackage[utf8]{inputenc}
\usepackage{array}
\usepackage{caption}
\usepackage{booktabs}
\usepackage{subcaption}
\usepackage{tabularx}
\usepackage{enumitem}
\usepackage{multirow}
\usepackage[hidelinks]{hyperref}
\usepackage{xurl}
\usepackage{xr} 
\usepackage[numbers]{natbib}
\def\BibTeX{{\rm B\kern-.05em{\sc i\kern-.025em b}\kern-.08em
    T\kern-.1667em\lower.7ex\hbox{E}\kern-.125emX}}
\begin{document}

\title{When Adaptation Fails: A Gradient-Based Diagnosis of Collapsed Gating in Vision-Language Prompt Learning}

\author{\IEEEauthorblockN{1\textsuperscript{st} Yunxuan Fang}
\IEEEauthorblockA{
\textit{Beihang University}\\
Beijing, China \\
micro@buaa.edu.cn}
\and
\IEEEauthorblockN{2\textsuperscript{nd} Ziwei Zhang}
\IEEEauthorblockA{
\textit{Beihang University}\\
Beijing, China \\
zwzhang@buaa.edu.cn}
\and
\IEEEauthorblockN{3\textsuperscript{rd} Xinhe Wang}
\IEEEauthorblockA{
\textit{Beihang University}\\
Beijing, China \\
xinhe@buaa.edu.cn}
}

\IEEEaftertitletext{\vspace{0.35\baselineskip}\begin{center}\footnotesize\copyright~2026 IEEE\end{center}}

\maketitle

\begin{abstract}
Adaptive prompting mechanisms have been proposed to enhance vision-language models by dynamically tailoring prompts to inputs. However, in frozen few-shot prompt learning with CLIP-style backbones, we systematically observe that adaptive gates and prompt-selection modules often collapse: they produce nearly constant outputs, contribute negligible gradient signals, and frequently fail to outperform fixed prompts.
To further explore this issue, we present a systematic diagnostic study to uncover the underlying causes and conditions of adaptation failure. Through controlled experiments across datasets and multiple prompt learning architectures, we identify two recurring failure modes: gradient magnitude imbalance and gate degradation.   
Our findings invite a re-examination of indiscriminately adding architectural complexity in parameter-efficient learning and clarify when prompt-level adaptive gating is, and is not, effective in this regime.
\end{abstract}

\begin{IEEEkeywords}
VLM, prompt learning, adaptive gating, gradient analysis, parameter-efficient learning
\end{IEEEkeywords}

\section{Introduction}
Large-scale vision-language models (VLMs) such as CLIP \cite{Radford2021LearningTV} exhibit remarkable zero-shot generalization capabilities, yet they often lack task-specific adaptability. Prompt learning has emerged as a leading parameter-efficient fine-tuning (PEFT) paradigm to address this gap~\cite{jia2022visual}, evolving from fixed prompts \cite{zhou2022learning} to conditional \cite{zhou2022conditional} and multi-modal variants \cite{khattakMaPLe}.
A natural extension is \textit{adaptive prompting}, which employs dynamic gating mechanisms to control prompt length or insertion depth per input. While the premise of such flexibility is appealing, our preliminary empirical evaluation reveals a stark divergence from this promise: adaptive components frequently collapse into near-constant outputs, contributing negligible gradient signal and failing to consistently outperform non-adaptive baselines. This raises a critical question: \textit{do these mechanisms truly adapt, or do they degenerate into static configurations?}

Motivated by this gap, we conduct a systematic diagnostic study to understand why and when adaptive prompting fails under frozen prompt-only tuning. Rather than proposing another performance-driven architecture, we use an adaptive bidirectional prompt learning framework (AdaptiveBiMaPLe) as a controlled testbed for tracing the path from optimization dynamics to functional collapse. 
Across multiple datasets (ImageNet~\cite{deng2009imagenet}, Caltech101~\cite{fei2004learning}, EuroSAT~\cite{helber2019eurosat}) and prompt learning frameworks (MaPLe~\cite{khattakMaPLe}, CoOp~\cite{zhou2022learning}, CoCoOp~\cite{zhou2022conditional}), we identify two recurring failure modes:
\begin{itemize}
\item \textbf{Gradient magnitude imbalance:} Gate parameters consistently receive gradients that are 2–3 orders of magnitude smaller than those of prompt parameters, creating an optimization barrier that prevents meaningful adaptation.
\item \textbf{Gate degradation:} Both length and depth gates converge to stable, near-constant values, showing no input-dependent variation and rendering the adaptive mechanism functionally equivalent to fixed prompts.
\end{itemize}

We further demonstrate that the occasional performance gain observed on small datasets (e.g., EuroSAT) is largely attributable to a parameter-count buffering effect, wherein the additional parameters of the gating module act as a regularization buffer rather than to genuine adaptive behavior. 
Our contributions are summarized as follows:
\begin{itemize}
\item We systematically analyze the failure modes of adaptive prompting across multiple CLIP-based prompt learning frameworks, identifying \textbf{gradient magnitude imbalance} and \textbf{gate degradation} as root causes.
\item We introduce a diagnostic procedure that links optimization failure to functional collapse and validate its generality through cross-model experiments.
\item We provide diagnostic criteria and design implications for evaluating adaptive prompting under frozen prompt-only tuning, including when additional architectural complexity is unlikely to provide meaningful benefit.
\end{itemize}

\section{Related Work}

\textbf{Prompt Learning for Vision-Language Models}: Prompt learning adapts pretrained VLMs by optimizing a small set of learnable tokens, avoiding full fine-tuning. CoOp \cite{zhou2022learning} pioneered this for CLIP by learning continuous context vectors, while CoCoOp \cite{zhou2022conditional} introduced image-conditioned prompts for better generalization. 
The paradigm was extended to multi-modal prompting with MaPLe \cite{khattakMaPLe}, which couples visual and textual prompts across transformer layers. More recent PEFT variants such as CLIP-Adapter~\cite{gao2024clip}, CLIPFit~\cite{li2024vision}, and DoPL~\cite{guo2025parameter} instead adapt features or backbone-affecting parameters, and are therefore structurally different from the frozen prompt-only gating regime studied here.

\textbf{Adaptive Gating Mechanisms}: Gating mechanisms dynamically control information flow in neural networks, from LSTMs \cite{hochreiter1997long} to modern transformers~\cite{vaswani2017attention}. Recent works explore adaptive gates for token pruning \cite{ye2025atp}, depth skipping, and mixture-of-experts models \cite{li2023adaptive}. In prompt learning, adaptive gating is an emerging idea intended to dynamically adjust prompt structure~\cite{zou2025prompthash, wang2025degap}. However, as we show in this paper, under frozen prompt-only tuning these gates often collapse, rendering them functionally close to fixed prompts in practice.

\textbf{Diagnostic Studies in Deep Learning}: Our work aligns with a growing body of literature that critically examines the internal mechanics of deep learning models. Previous diagnostic studies have revealed issues like attention collapse \cite{zhai2023stabilizing, noci2022signal}, gradient starvation \cite{pezeshki2021gradient}, and mode collapse in generative models~\cite{kossale2022mode}. We extend this line of inquiry to adaptive prompting in VLMs through a gradient-based diagnosis of gating failure.

\section{Experimental Setup and Testbed Model}

This section outlines our diagnostic testbed for analyzing adaptive gating failures. Rather than pursuing performance gains, we introduce AdaptiveBiMaPLe primarily to investigate the optimization dynamics and representational behavior of adaptive mechanisms.

\subsection{Base Architecture: Revisiting MaPLe}

MaPLe (Multi-modal Prompt Learning) \cite{khattakMaPLe} adapts both vision and language branches of CLIP through deep prompting across transformer layers. For the language branch at layer $d$, learnable prompts $P_t^{(d)} \in \mathbb{R}^{N \times D_t}$ are concatenated with word embeddings:
\[
[P_t^{(d)}, W^{(d)}] = \mathcal{L}^{(d)}([P_t^{(d-1)}, W^{(d-1)}]),
\]
where $\mathcal{L}^{(d)}$ denotes the $d$-th transformer layer, $W^{(d)}$ are the word embeddings, and $N$ is the prompt length.
Similarly, visual prompts $P_v^{(d)} \in \mathbb{R}^{N \times D_v}$ are introduced in the vision encoder. The key innovation is vision-language coupling: $P_v^{(d)} = f_{L\rightarrow V}^{(d)}(P_t^{(d)})$, where $f_{L\rightarrow V}^{(d)}(\cdot)$ is a linear projection that ensures mutual synergy between modalities.

\subsection{Bidirectional Extension: BiMaPLe}
We extend MaPLe with bidirectional cross-modal coupling. For each layer $d$, two lightweight MLPs map representations between modalities:
\[
\tilde{P}_v^{(d)} = f_{L\rightarrow V}^{(d)}(P_t^{(d)}), \quad 
\tilde{P}_t^{(d)} = f_{V\rightarrow L}^{(d)}(P_v^{(d)}).
\]
The coupled prompts are obtained through learnable fusion:
\[
P_t^{(d)*} = \alpha_d P_t^{(d)} + (1-\alpha_d)\tilde{P}_t^{(d)},\]
\[
P_v^{(d)*} = \beta_d P_v^{(d)} + (1-\beta_d)\tilde{P}_v^{(d)},
\]
where $\alpha_d, \beta_d \in (0,1)$ are trainable coupling coefficients. A cycle consistency loss $\mathcal{L}_\mathrm{cyc}$ regularizes the mappings to be approximately invertible. The detailed architecture and formulations are provided in Sec~\ref{sec:BiMaPLe} of the Appendix.

\subsection{Adaptive Gating: AdaptiveBiMaPLe}
Based on BiMaPLe, AdaptiveBiMaPLe further introduces two levels of adaptive control.

\textbf{Length Gating.} For each token $p_i^{(d)}$ at depth $d$, a learnable scalar gate $g_i^{(d)}$ controls its activation:
\[
\tilde{p}_i^{(d)} = \sigma(g_i^{(d)}) \cdot p_i^{(d)}, \quad
\widetilde{P}_t^{(d)} = \{\tilde{p}_1^{(d)}, \dots, \tilde{p}_{N_{\max}}^{(d)}\},
\]
where $\sigma(\cdot)$ denotes the sigmoid function. The effective prompt length is measured as $L_{\mathrm{eff}}^{(d)} = \sum_{i=1}^{N_{\max}} \sigma(g_i^{(d)})$.

\textbf{Depth Gating.} For layer-wise insertion control, learnable parameters $g_\mathrm{depth}^{(d)}$ determine insertion strength: $w_d = \sigma(g_\mathrm{depth}^{(d)})$.
Prompts are softly inserted with weight $w_d$ during training and we threshold $\sigma(g_\mathrm{depth}^{(d)})>0.5$ during inference for hard insertion decisions~\cite{he2018soft}.

Two auxiliary losses are used to stabilize the training:
\begin{itemize}
    \item Sparsity regularization to encourage compact prompts:
    \[\mathcal{L}_\mathrm{sparse} =  \sum \nolimits_{d}\sum \nolimits_{i} \sigma(g_i^{(d)}).\]
    \item Depth smoothness regularization to enforce gradual depth variation:
    \[\mathcal{L}_\mathrm{smooth} =  \sum \nolimits_{d} \left|g_\mathrm{depth}^{(d+1)} - g_\mathrm{depth}^{(d)}\right|.\]
\end{itemize}

The overall objective combines classification loss with regularization terms:
\[
\mathcal{L}_\mathrm{total} = \mathcal{L}_\mathrm{cls} + \mathcal{R}(\Theta),
\]
where \(\mathcal{R}(\Theta) = \lambda_\mathrm{cyc} \mathcal{L}_\mathrm{cyc} + \lambda_\mathrm{sparse} \mathcal{L}_\mathrm{sparse} + \lambda_\mathrm{smooth} \mathcal{L}_\mathrm{smooth}\) denotes the composite regularization term.
More details are provided in Sec~\ref{sec:AdaptiveBiMaPLe} of the Appendix.

\subsection{Cross-Model Adaptations: CoOp and CoCoOp}

To ensure that our empirical evaluations reflect a general phenomenon, we adopt two additional adaptive gating mechanisms in representative CLIP-based prompt learning frameworks: CoOp \cite{zhou2022learning} and CoCoOp \cite{zhou2022conditional}.

\textbf{CoOp-gating:} For context tokens $C = [C_1, C_2, ..., C_M]$, we introduced per-token scalar gates: $\tilde{C_i} = \sigma(g_i) \cdot C_i$, where $g_i$ are learnable gate parameters.
The effective context length is measured as $L_{\mathrm{eff}} = \sum_{i=1}^M \sigma(g_i)$.

\textbf{CoCoOp-gating:} For CoCoOp's instance-conditioned prompts $C(x) = [C_1(x), C_2(x), ..., C_M(x)]$, we implemented input-dependent gating:
\[
\tilde{C_i}(x) = \sigma(w_i^\top v(x) + b_i) \cdot C_i(x),
\]
where $v(x)$ denotes the normalized image-encoder features, and $w_i, b_i$ are learnable parameters. CoCoOp's meta-network separately generates the instance-conditioned prompt bias.

\subsection{Implementation Details}

We follow MaPLe's \cite{khattakMaPLe} experimental protocol across three datasets: ImageNet \cite{deng2009imagenet}, Caltech101 \cite{fei2004learning}, and EuroSAT \cite{helber2019eurosat}. All experiments use frozen CLIP-ViT-B/16 backbone, trained on base classes (16-shot) and evaluated on both base and novel classes. More details are provided in Sec~\ref{sec:Implementation_Details} of the Appendix.

\section{Diagnostic Experiments and Analysis}

To understand why adaptive prompting mechanisms fail to realize meaningful flexibility, we conduct a sequence of targeted diagnostic experiments. Each experiment is designed to answer a specific question about gate behavior, optimization dynamics, or representational properties. 

\subsection{Overall Performance}

We begin by evaluating AdaptiveBiMaPLe on ImageNet, Caltech101, and EuroSAT, alongside MaPLe and BiMaPLe. Table~\ref{table: overall acc} summarizes the results. On ImageNet and Caltech101, AdaptiveBiMaPLe consistently underperforms non-adaptive baselines, even when using a 50× larger learning rate for gate parameters. On EuroSAT, however, the adaptive variant achieves higher accuracy. 
These observations motivate the following in-depth analyses.

\begin{table}[t]
\centering
\small
\captionsetup[subtable]{skip=5pt}
\begin{subtable}{0.48\textwidth}
  \centering
  \caption{ImageNet}
  \begin{tabular}{lccc}
  \toprule
  Model & Base Acc & Novel Acc & H-Mean \\
  \midrule
  MaPLe & \textbf{77.07} & 70.43 & 73.60 \\
  BiMaPLe & 76.73 & \textbf{71.00} & \textbf{73.76} \\
  AdaptiveBiMaPLe & 76.23 & 70.20 & 73.09 \\
  AdaptiveBiMaPLe* & 75.86 & 69.80 & 72.70 \\
  \bottomrule
  \end{tabular}
\end{subtable}
\hfill
\begin{subtable}{0.48\textwidth}
  \centering
  \caption{Caltech101}
  \begin{tabular}{lccc}
  \toprule
  Model & Base Acc & Novel Acc & H-Mean \\
  \midrule
  MaPLe & \textbf{98.23} & \textbf{95.10} & \textbf{96.64} \\
  BiMaPLe & 98.00 & 94.80 & 96.37 \\
  AdaptiveBiMaPLe & 97.70 & 93.57 & 95.59 \\
  AdaptiveBiMaPLe* & 97.67 & 93.60 & 95.59 \\
  \bottomrule
  \end{tabular}
\end{subtable}
\hfill
\begin{subtable}{0.48\textwidth}
  \centering
  \caption{EuroSAT}
  \begin{tabular}{lccc}
  \toprule
  Model & Base Acc & Novel Acc & H-Mean \\
  \midrule
  MaPLe & 83.87 & 57.65 & 68.33 \\
  BiMaPLe & 82.68 & 63.08 & 71.56 \\
  AdaptiveBiMaPLe & \textbf{85.73} & 68.48 & 76.14 \\
  AdaptiveBiMaPLe* & \textbf{85.73} & \textbf{68.80} & \textbf{76.33} \\
  \bottomrule
  \end{tabular}
\end{subtable}
\caption{The overall model performance. \textbf{Adaptive gating fails to improve performance on ImageNet or Caltech101.} Despite increased flexibility, AdaptiveBiMaPLe underperforms BiMaPLe and MaPLe on ImageNet and Caltech101. 
All values are harmonic means over 3 random seeds.}
\label{table: overall acc}
\end{table}

\subsection{Diagnosing Optimization Failure}

\begin{figure}[htbp]
\centering
\includegraphics[width=0.48\textwidth]{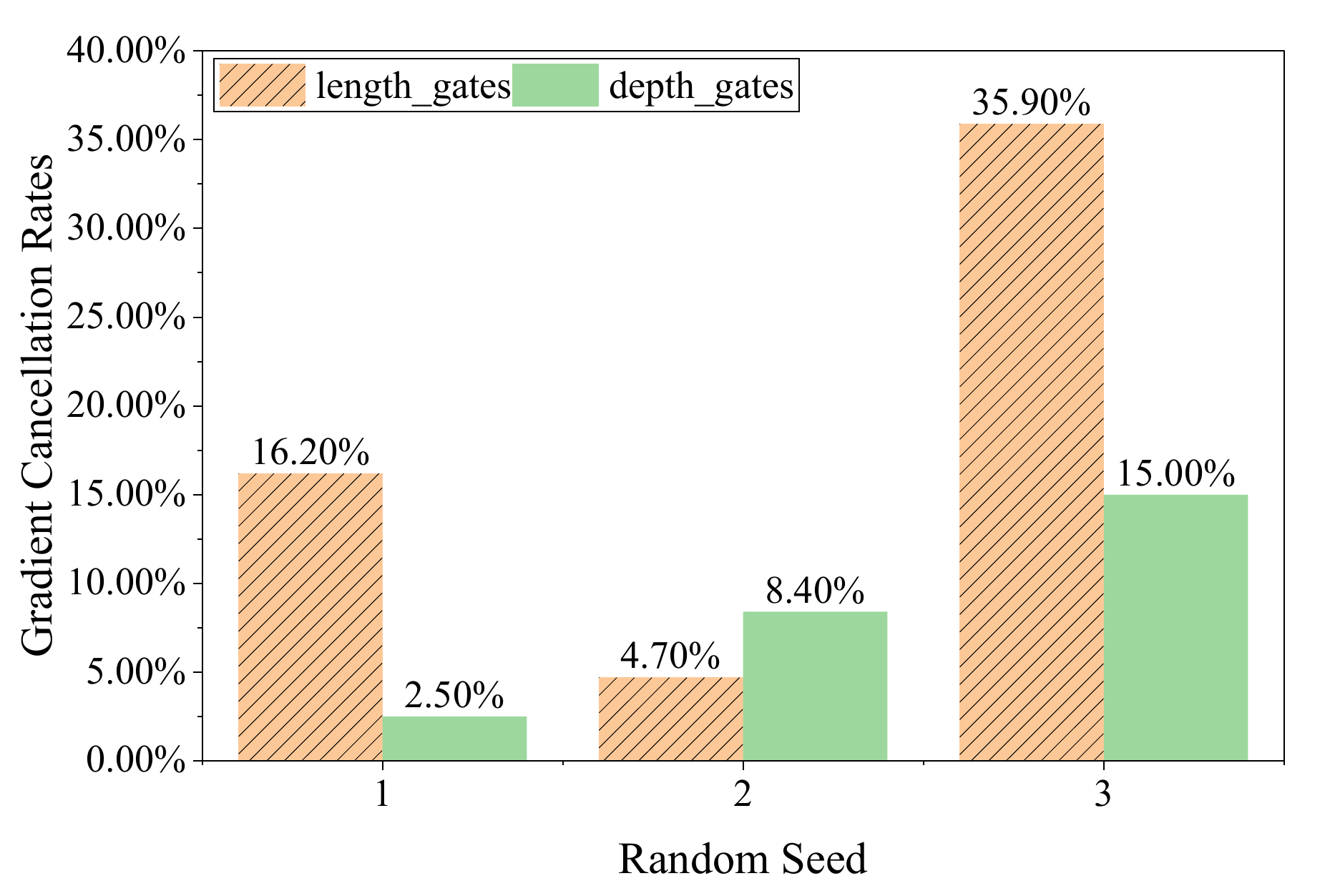}
\caption{\textbf{Gradient Cancellation Rates.}
Cancellation rates for length and depth gate parameters on ImageNet. Each bar represents averages over three random seeds.}
\label{fig:cancellation rate}
\end{figure}

\subsubsection{Gradient Flow Analysis}

\begin{figure*}[htbp]
\centering
\includegraphics[width=0.98\textwidth]{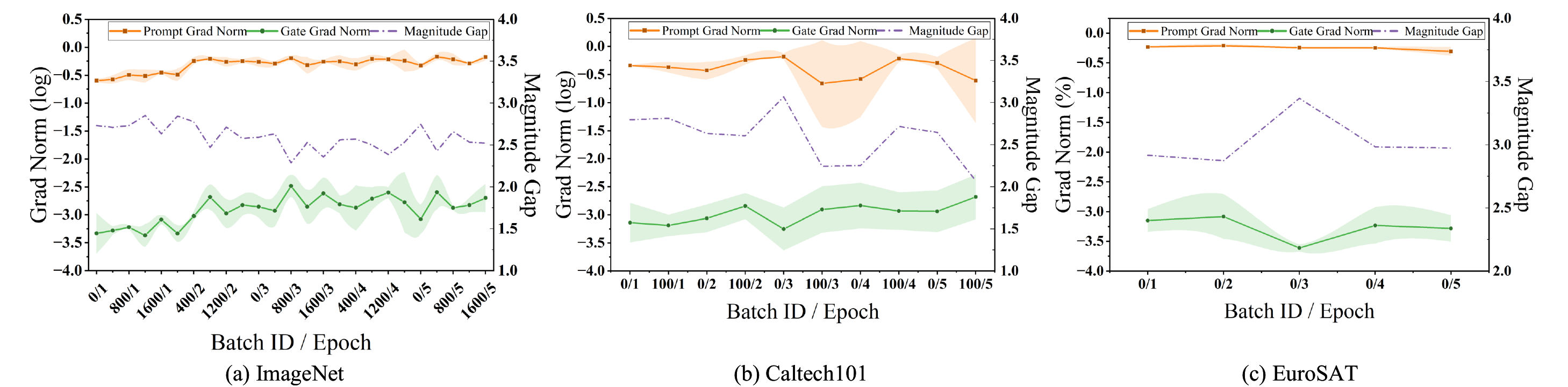}
\caption{\textbf{Gradient Norm Comparison.}
Gradient norms of prompt parameters and gate parameters measured across training iterations.
Shaded regions indicate variation across random seeds. Gate parameters receive gradients 2-3 orders smaller than prompt parameters across all datasets.}
\label{fig:gradients}
\end{figure*}

To determine whether adaptive gates receive sufficient optimization signal, we monitor the gradient norms of prompt parameters and gate parameters throughout training. Figure~\ref{fig:gradients} shows that gate gradients remain 2-3 orders of magnitude smaller than prompt gradients across datasets and seeds. 
This severe imbalance persists for AdaptiveBiMaPLe*, which uses 50× higher learning rates for gates, indicating fundamental scaling issues rather than mere step size problems (Table~\ref{table: LR scaling}).

In Figure~\ref{fig:cancellation rate}, we further investigate the gradient cancellation rates. The results reveal directional instability, i.e., gate gradients show significant variability across seeds and epochs, suggesting they are affected by noise in the optimization.

\begin{table}[bt]
\centering
\small
\begin{tabular}{lccc}
\toprule
Gate Strategy & ImageNet & Caltech101 & EuroSAT \\
\midrule
Fixed (All-on) & 75.31 & 96.49 & 79.41 \\
Random & 74.18 & 96.17 & 73.82 \\
Per-Layer & 75.31 & 96.49 & 79.41 \\
\specialrule{0.06em}{0.25em}{0.25em}
Adaptive (Per-Token) & 75.31 & 96.49 & 79.41 \\
\bottomrule
\end{tabular}
\caption{\textbf{Performance under Different Gating Strategies.} All gating strategies achieve nearly identical performance.}
\label{table: different gate strategy}
\end{table}

\begin{table}[b]
\centering
\small
\begin{tabular}{lcccc}
\toprule
Model & Alignment & Clustering & Separation & H-Mean \\
\midrule
MaPLe & 0.3179 & 0.5622 & 0.3119 & 73.60 \\
BiMaPLe & 0.3208 & 0.5663 & 0.3248 & 73.76 \\
Adaptive & 0.3113 & 0.5535 & 0.3237 & 73.09 \\
\bottomrule
\end{tabular}
\caption{\textbf{Representation Quality Metrics.} The adaptive module neither significantly enhances nor compromises foundational CLIP representations.}
\label{table: representation quality}
\end{table}

\begin{table*}[htbp]
\centering
\small
\begin{tabular}{lccc}
\toprule
Model & Prompt Grad Norm &  Gate Grad Norm & Magnitude Gap \\
\midrule
AdaptiveBiMaPLe & $5.11{\times}10^{-1} \pm 2.56{\times}10^{-2}$ & $1.59{\times}10^{-3} \pm 8.32{\times}10^{-5}$ & $2.60 \pm 1.87{\times}10^{-2}$ \\
AdaptiveBiMaPLe* & $5.04{\times}10^{-1} \pm 1.37{\times}10^{-2}$ & $2.56{\times}10^{-3} \pm 2.24{\times}10^{-4}$ & $2.43 \pm 1.73{\times}10^{-2}$ \\
\bottomrule
\end{tabular}
\caption{\textbf{Gradient norms with different learning rates.} The results are from ImageNet while other datasets show similar patterns. Values represent arithmetic mean ± standard deviation computed over 3 random seeds.}
\label{table: LR scaling}
\end{table*}

\subsubsection{Gate Behavior Analysis}

\begin{figure}[htbp]
\centering
\includegraphics[width=0.48\textwidth]{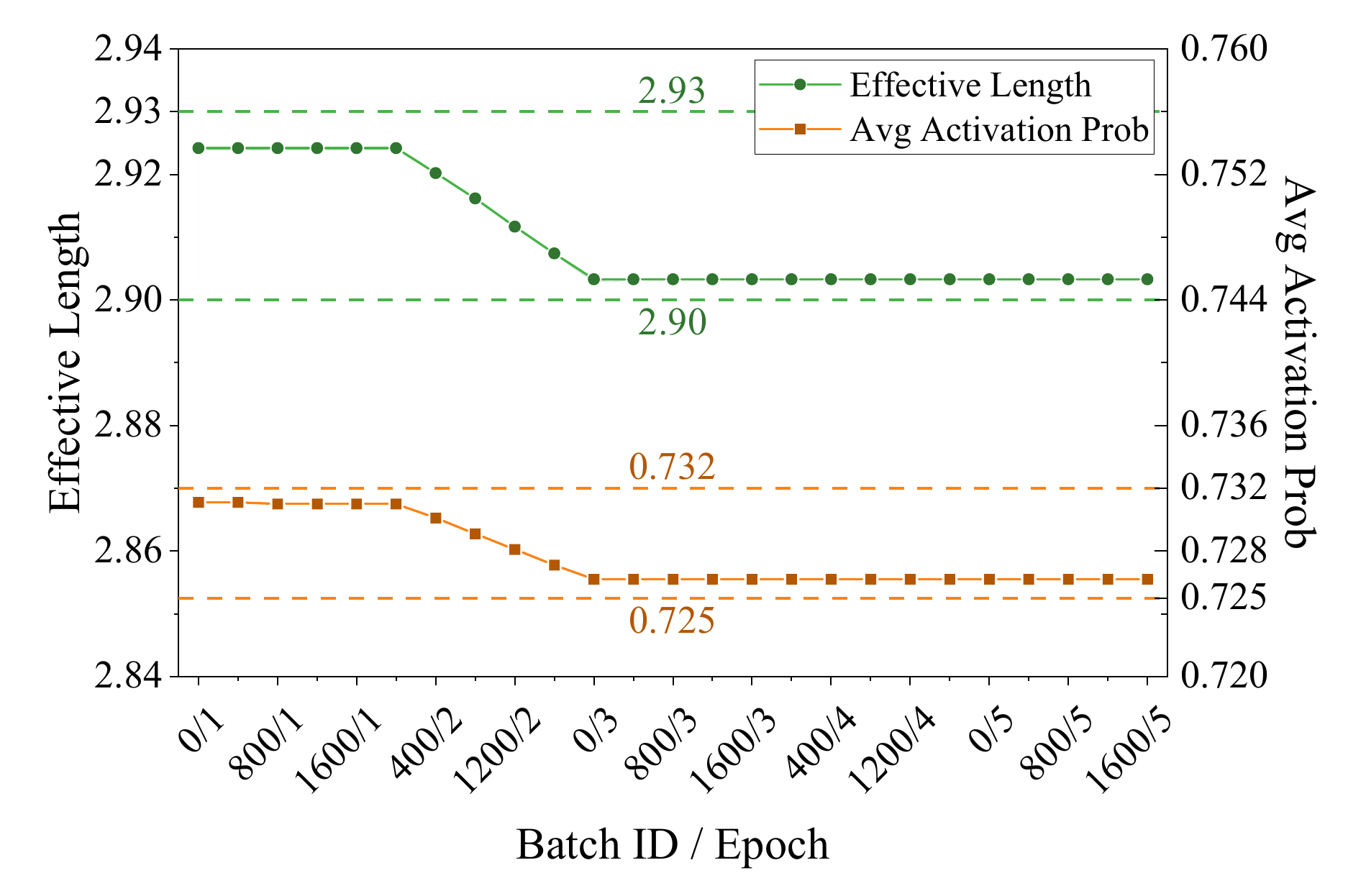}
\caption{\textbf{The effective prompt lengths and depth activation probabilities during training.} Both quantities remain nearly constant throughout training.}
\label{fig:gate_behavior}
\end{figure}

We further investigate whether gates exhibit meaningful adaptive behavior during training by tracking the effective prompt lengths and depth activation probabilities with respect to the training process. 
As shown in Figure~\ref{fig:gate_behavior}, both measures remain nearly constant during training, indicating limited changes in the learned gate strengths.

We further assess whether this limited gate variation is prevalent across gating strategies. As shown in Table~\ref{table: different gate strategy}, similar results are observed for different gating strategies, as fixed, adaptive, and randomized gates achieve nearly identical accuracy, demonstrating possible redundancy of the adaptive mechanism.

The stable accuracy alongside nearly unchanged gate statistics is consistent with optimization being dominated by CLIP's frozen features and MaPLe-style prompts. In this setting, the added gates provide limited observed benefit over fixed-prompt variants.

\subsection{Additional Experiments and Further Analyses}

\subsubsection{Representation Quality Analysis}
\label{sec: representation quality}

To assess whether adaptive gates affect fundamental representation quality, we evaluate image-text alignment, clustering quality (Silhouette Score), and inter-class separation. As shown in Table~\ref{table: representation quality}, all models show comparable performance on these metrics, indicating that the adaptive module neither enhances nor compromises CLIP's foundational representations. The observed performance differences likely stem from optimization efficiency in prompt space rather than representation capacity.

\subsubsection{Ablation Studies and Sensitivity Analysis}

To test the robustness of our findings, we perform ablations on regularization weights ($\lambda_{\mathrm{sparse}}$, $\lambda_{\mathrm{smooth}}$, $\lambda_{\mathrm{cyc}}$) and gating configurations (full, length-only, depth-only). Across all settings, accuracy varies by less than 0.5\%, indicating insensitivity to hyperparameters or architectural choices, supporting the interpretation that the adaptive components remain largely inactive. 

To further validate that the gate collapse is not due to suboptimal hyperparameter settings but a structural issue, we conducted an extended series of optimization strategy ablations. These included systematic attempts to balance gradients and to prevent gate saturation, as detailed in Appendix~\ref{sec:gradient_balancing} and \ref{sec:revive_gating}. None of these strategies successfully restored meaningful gate adaptation or improved performance beyond the fixed-prompt baselines, reinforcing that the collapse stems from structural gradient attenuation rather than insufficient tuning.

Details of the standard ablation studies are provided in Section~\ref{sec:Ablation_Studies} of the Appendix.

\subsubsection{Computational Efficiency}
AdaptiveBiMaPLe increases parameter count by 38\% (13.8M vs. 10.1M) and FLOPs by 27\% compared to BiMaPLe, yet does not lead to accuracy improvements. 
The additional complexity provides no discernible benefit.

\subsubsection{The EuroSAT Dataset Analysis}
\label{sec:dataset_behavior}

From Table~\ref{table: overall acc}, the results on EuroSAT dataset show a different pattern from the other two, where AdaptiveBiMaPLe indeed improves the performance, particularly for the novel classes.
To investigate this inconsistency, we test hypotheses through controlled variants, including removing gates (ParamMatched), keeping gates frozen (AlwaysFrozen), and applying regularization (ExplicitReg). 

\begin{table}[b]
\centering
\setlength{\tabcolsep}{3pt}
\small
\begin{tabularx}{0.48\textwidth}{@{} l >{\raggedright\arraybackslash}X c @{}}
\toprule
Variant & Description & H-Mean \\
\midrule
ParamMatched & Remove gates, add equivalent parameters & 80.60 \\
AlwaysFrozen & Keep gates but freeze from start & 74.29 \\
ExplicitReg & Apply dropout/weight decay & 80.77 \\
\midrule
AdaptiveBiMaPLe & Original adaptive model & 76.14 \\
\bottomrule
\end{tabularx}
\caption{\textbf{EuroSAT Ablation Variants.} ParamMatched and ExplicitReg achieve the highest performance. Values represent harmonic mean over 3 random seeds.}
\label{table: eurosat exception}
\end{table}

The results are shown in Table~\ref{table: eurosat exception}. ParamMatched and ExplicitReg achieve comparable or higher performance despite removing genuine adaptive behavior. These findings suggest that the EuroSAT gains are not uniquely attributable to dynamic gating. Instead, they are more consistent with parameter-count buffering and regularization effects in small-data regimes. While domain-specific factors may also play a role, our controlled variants indicate that adaptive gating itself is not necessary to obtain the observed improvement.

\section{Cross-Model Validation: On the Boundary of Adaptive Gating}

The previous experiments focus on the AdaptiveBiMaPLe framework, where adaptive gating is applied to a multimodal deep prompting architecture. To determine whether the observed failure modes are specific to this architecture or reflect a broader limitation, we extend our analysis to two additional prompting frameworks based on CLIP: CoOp \cite{zhou2022learning} and CoCoOp \cite{zhou2022conditional}. These models differ substantially in structure, where CoOp uses fixed learnable text prompts, while CoCoOp conditions prompts on image features.

\subsection{Performance Comparison Across Frameworks}

We first compare the original and gated variants of CoOp and CoCoOp. Table~\ref{table:cross_model_performance} summarizes the results. Across both ImageNet and EuroSAT, gated versions do not provide consistent performance improvement over the originals. 

\begin{table}[htbp]
\centering
\small
\setlength{\tabcolsep}{3.5pt}
\begin{tabular}{@{}lcccc@{}}
\toprule
Framework & Variant & ImageNet & Caltech101 & EuroSAT \\
\midrule
\multirow{2}{*}{CoOp} & Original & $73.22_{\pm0.38}$ & \textbf{$96.14_{\pm0.55}$} & \textbf{$69.67_{\pm10.99}$} \\
 & Gating & \textbf{$73.38_{\pm0.31}$} & $95.74_{\pm0.31}$ & $69.00_{\pm3.72}$ \\
\specialrule{0.06em}{0.25em}{0.25em}
\multirow{2}{*}{CoCoOp} & Original & \textbf{$73.02_{\pm0.22}$} & \textbf{$95.59_{\pm0.28}$} & \textbf{$72.98_{\pm4.76}$} \\  
 & Gating & $72.98_{\pm0.21}$ & $95.47_{\pm0.09}$  & $71.07_{\pm8.71}$ \\
\bottomrule
\end{tabular}
\caption{\textbf{Cross-Model Performance.} Adaptive gating fails to provide consistent benefits over their original counterparts. Values are harmonic means over three random seeds.}
\label{table:cross_model_performance}
\end{table}

\subsection{Diagnosing Results}
\subsubsection{Gradient Analysis}

To examine whether similar optimization challenges arise, we measure gradient norms for gate parameters in CoOp and CoCoOp. As shown in Table~\ref{table:cross_model_gradients}, gate gradients remain 2-6 orders of magnitude smaller than prompt gradients across both frameworks, and CoCoOp exhibits the most severe attenuation due to its deeper conditioning network. These measurements show that gradient starvation is not unique to AdaptiveBiMaPLe, but occurs consistently across architectures that incorporate adaptive gating.

\begin{table}[htbp]
\centering
\small
\begin{tabular}{lcc}
\toprule
Framework & Gate Gradient Norm & Performance Gap \\
\midrule
MaPLe & $1.59\times10^{-3}$ &  $-0.67$ \\
CoOp & $3.60\times10^{-3}$ &   $+0.16$ \\
CoCoOp & $7.47\times10^{-6}$ & $-0.04$ \\
\bottomrule  
\end{tabular}
\caption{\textbf{Gate Gradient Norms Across Frameworks.} All adaptive variants exhibit severely attenuated gate gradients. The values are averages on ImageNet over 3 random seeds.}
\label{table:cross_model_gradients}
\end{table}

\subsubsection{Gate Behavior}

We further examine the gate statistics in CoOp and CoCoOp. For CoOp, the logged soft effective length is approximately 7.96 across 16 context tokens (mean activation 0.4975). For conditional CoCoOp, the corresponding value is approximately 1.995 across four tokens (mean activation 0.49875), with logged gate gradients on the order of $10^{-6}$. Both models thus show average gate activations close to 0.5.

\subsection{Implications for Adaptive Prompting}

The consistent failure of adaptive gating across three distinct prompt learning frameworks suggests that the problem is not specific to AdaptiveBiMaPLe, but recurs across multiple prompt-level gating designs under frozen few-shot prompt learning. The shared symptoms—gradient magnitude imbalance, gate convergence to trivial patterns, and limited performance benefit—point to a structural optimization difficulty in this regime.

\section{Theoretical Analysis and Discussion}

Our empirical findings point to two consistent failure modes in adaptive prompting: gradient imbalance and gate collapse. To understand why these failures occur, we bridge our observations with theoretical insights, aiming to explain the underlying mechanisms and derive principles for future designs.

\subsection{Theoretical Analysis of Gradient Attenuation}

We examine how the mathematical structure of sigmoid gating can contribute to the observed gradient imbalance. 
Consider a learnable gate vector $g_i \in \mathbb{R}^{N_{\max}}$ modulating prompt tokens $p_i$ via $\tilde{p}_i = \sigma(g_i) \cdot p_i$,
where $\sigma(\cdot)$ denotes the sigmoid function. The gradient of the loss $\mathcal{L}$ with respect to $g_i$ is:
\[
\frac{\partial \mathcal{L}}{\partial g_i}
= \frac{\partial \mathcal{L}}{\partial \tilde{p}_i}
  \cdot \frac{\partial \tilde{p}_i}{\partial g_i}
= \frac{\partial \mathcal{L}}{\partial \tilde{p}_i}
  \cdot \sigma(g_i)(1-\sigma(g_i)) \, p_i.
\]

The term $\sigma(g_i)(1-\sigma(g_i))$ is bounded by 0.25 and vanishes as $\sigma(g_i)$ approaches 0 or 1. This creates a structural gradient attenuation: even if the prompt gradient $\partial \mathcal{L} / \partial \tilde{p}_i$ is large, the gate gradient is inherently scaled down. If gate logits move toward extreme values, this factor can further weaken their gradients.

\subsection{Implications for Adaptive Mechanism Design}

Our analysis suggests that simple sigmoidal prompt gating is difficult to optimize in few-shot prompt learning with frozen backbones.
The consistent gradient starvation implies that adaptive components require either alternative activation functions with non-vanishing gradients, or separate optimization pathways to avoid competition with prompt parameters, or meta-learned gating updates that circumvent direct gradient-based updates.
Moreover, the fact that explicit regularization outperforms unactivated adaptive gates (especially on small datasets) suggests that robustness may be better achieved through structured regularization rather than added architectural complexity. This insight invites a reconsideration of the prevailing "more gates, more flexibility" design philosophy.

\subsection{Limitations and Future Directions}

Our study focuses on CLIP-ViT-B/16 and classification tasks, while extension to other architectures and domains remains valuable. 
We also see promise in per-sample gating mechanisms, though their computational cost must be carefully weighed against potential gains. Ultimately, our diagnostic approach provides a template for evaluating future adaptive designs: if gate gradients remain orders of magnitude smaller than those of the tuned parameters, collapse is likely imminent.
Our conclusions are specific to prompt-level gating under frozen backbones and should not be conflated with gating in architectures such as mixture-of-experts, where gates control fully trainable experts through much stronger gradient pathways; see supplementary material for discussion.

\subsection{Broader Impact}

Beyond adaptive prompting, our study highlights a general tension in parameter-efficient learning: flexibility often comes at the cost of optimization stability. Architectural innovations must be paired with corresponding advances in optimization strategies to ensure balanced gradient flow. Moreover, we advocate for a diagnostic-first approach in model development: monitoring internal dynamics like gradient norms and activation patterns can reveal failure modes long before they manifest in performance metrics.

\section{Conclusion}

In this paper, we investigated adaptive gating collapse in vision-language prompt learning through a systematic diagnostic study. Focusing on frozen few-shot prompt learning with CLIP-style backbones, we identified two recurring failure modes: gradient magnitude imbalance and gate degradation. Our results show that prompt-level sigmoidal gating is difficult to optimize in this regime and can easily become functionally inert. These findings motivate a more critical evaluation of the complexity-flexibility trade-off and provide a practical diagnostic lens for future adaptive vision-language designs.

\bibliographystyle{IEEEtran}
\bibliography{ICME_Template/ref_short}

\newpage
\appendix

\subsection{Supplementary Material Overview}

This supplementary material provides supporting details for the camera-ready version of the paper. Its purpose is not to introduce a substantially different manuscript, but to document the technical details, additional ablations, and reviewer-requested clarifications that could not be fully included in the 6-page main paper. In particular, this document complements the main paper in three ways: 
(1) detailed formulations of BiMaPLe and AdaptiveBiMaPLe, 
(2) implementation details and extended ablations, and 
(3) additional clarifications on EuroSAT, gradient variance, long-horizon training, and the distinction between prompt-level gating and MoE-style gating.

\subsection{Detailed Method Formulation}

\subsubsection{BiMaPLe: Bidirectional Cross-Modal Prompt Coupling}
\label{sec:BiMaPLe}

BiMaPLe extends MaPLe by introducing bidirectional coupling between textual and visual prompts. For each transformer layer $d$, we maintain textual prompts $P_t^{(d)} \in \mathbb{R}^{N \times D_t}$ and visual prompts $P_v^{(d)} \in \mathbb{R}^{N \times D_v}$. Two lightweight mapping networks translate prompts between modalities:
\[
\tilde{P}_v^{(d)} = f_{L\rightarrow V}^{(d)}(P_t^{(d)}), \qquad
\tilde{P}_t^{(d)} = f_{V\rightarrow L}^{(d)}(P_v^{(d)}).
\]

The coupled prompts are then obtained by learnable fusion:
\[
P_t^{(d)*} = \alpha_d P_t^{(d)} + (1-\alpha_d)\tilde{P}_t^{(d)},
\]
\[
P_v^{(d)*} = \beta_d P_v^{(d)} + (1-\beta_d)\tilde{P}_v^{(d)},
\]
where $\alpha_d,\beta_d \in (0,1)$ are trainable scalar coefficients.

To encourage consistency between the two projections, we apply a cycle-consistency regularizer:
\[
\mathcal{L}_{\mathrm{cyc}} = \frac{1}{D}\sum_d 
\left\|P_t^{(d)} - f_{V\rightarrow L}^{(d)}\!\left(f_{L\rightarrow V}^{(d)}(P_t^{(d)})\right)\right\|_2^2.
\]

This bidirectional design serves as the non-adaptive baseline from which AdaptiveBiMaPLe is constructed.

\subsubsection{AdaptiveBiMaPLe: Prompt Length and Depth Gating}
\label{sec:AdaptiveBiMaPLe}

AdaptiveBiMaPLe augments BiMaPLe with two gating mechanisms.

\paragraph{Length gating.}
For each token $p_i^{(d)}$ at layer $d$, we introduce a learnable scalar gate $g_i^{(d)}$:
\[
\tilde{p}_i^{(d)} = \sigma(g_i^{(d)}) \cdot p_i^{(d)}.
\]
The effective prompt length is measured as
\[
L_{\mathrm{eff}}^{(d)}=\sum_{i=1}^{N_{\max}}\sigma(g_i^{(d)}).
\]

\paragraph{Depth gating.}
For each insertion depth $d$, we define a learnable gate parameter $g_{\mathrm{depth}}^{(d)}$ and insertion weight
\[
w_d = \sigma(g_{\mathrm{depth}}^{(d)}).
\]
During training, prompts are inserted softly with weight $w_d$. During inference, we optionally threshold $\sigma(g_{\mathrm{depth}}^{(d)})>0.5$ to obtain a hard insertion decision. In practice, because gate values remain close to their initialization throughout training, the final behavior is largely insensitive to this threshold choice.

\paragraph{Regularization terms.}
The overall objective is
\[
\mathcal{L}_{\mathrm{total}}
=
\mathcal{L}_{\mathrm{cls}}
+
\lambda_{\mathrm{cyc}}\mathcal{L}_{\mathrm{cyc}}
+
\lambda_{\mathrm{sparse}}\mathcal{L}_{\mathrm{sparse}}
+
\lambda_{\mathrm{smooth}}\mathcal{L}_{\mathrm{smooth}},
\]
where
\[
\mathcal{L}_{\mathrm{sparse}} = \sum_d\sum_i \sigma(g_i^{(d)}),
\]
\[
\mathcal{L}_{\mathrm{smooth}} = \sum_d
\left|g_{\mathrm{depth}}^{(d+1)}-g_{\mathrm{depth}}^{(d)}\right|.
\]

\subsubsection{Cross-Model Gating Variants for CoOp and CoCoOp}

To test whether the observed failure mode is architecture-specific, we implemented prompt-level gating in CoOp and CoCoOp.

\paragraph{CoOp-gating.}
For learnable context tokens $C=[C_1,\dots,C_M]$, we introduce per-token scalar gates:
\[
\tilde{C}_i=\sigma(g_i)\cdot C_i,
\qquad
L_{\mathrm{eff}}=\sum_{i=1}^M \sigma(g_i).
\]

\paragraph{CoCoOp-gating.}
For instance-conditioned prompts $C(x)=[C_1(x),\dots,C_M(x)]$, we implement input-dependent gating:
\[
\tilde{C}_i(x)=\sigma(w_i^\top v(x)+b_i)\cdot C_i(x),
\]
where $v(x)$ denotes normalized image-encoder features; the CoCoOp meta-network separately supplies the instance-conditioned prompt bias.

These two variants allow us to test whether gradient attenuation and gate collapse persist beyond the AdaptiveBiMaPLe formulation.

\subsection{Experimental Details}
\label{sec:Implementation_Details}

\subsubsection{Datasets, Backbone, and Optimization}

We follow the standard few-shot prompt-learning protocol used in prior CLIP-based work. All experiments use a frozen CLIP ViT-B/16 backbone and are trained on base classes in the 16-shot setting, with evaluation on both base and novel classes.

We report results on three classification benchmarks of different scales and domains: ImageNet, Caltech101, and EuroSAT. Unless otherwise specified, optimization follows the MaPLe protocol. For AdaptiveBiMaPLe, gate parameters are initialized to 1.0, and the regularization weights are set to $\lambda_{\mathrm{sparse}}=0.001$, $\lambda_{\mathrm{smooth}}=0.0002$, and $\lambda_{\mathrm{cyc}}=0.1$.

All reported results are averaged over three independent random seeds. The main paper reports harmonic-mean accuracy across seeds, while gradient statistics are reported as arithmetic mean $\pm$ standard deviation across seeds.

\subsubsection{Training Horizon, Variance, and Reproducibility}

All methods in the main paper follow the standard training schedules used by the original prompt-learning baselines. In response to reviewer concerns about training horizon, we additionally extended representative runs up to 10 epochs. We consistently observed that gate gradients decay early and then remain strongly attenuated, while effective prompt length and depth activation statistics stay nearly constant throughout the extended horizon. This suggests that the observed collapse is not merely a short-training artifact.

Regarding seed variance, the main paper already reports all major results over three independent seeds. The variance in gradient cancellation rates reflects the fact that once gate gradients are 2--3 orders of magnitude smaller than prompt gradients, stochastic minibatch noise can significantly affect their effective update direction. In other words, the small absolute scale of gate gradients makes them directionally unstable even when the final gate activations still collapse to similar near-constant values.

Code is available at: \url{https://github.com/MiCrSYZ/when_adaptation_fails.git}.

\subsection{Additional Diagnostics and Ablations}

\subsubsection{Standard Ablations}
\label{sec:Ablation_Studies}

We perform two types of standard ablations on ImageNet.

\paragraph{Regularization weights.}
We vary $\lambda_{\mathrm{sparse}}$, $\lambda_{\mathrm{smooth}}$, and $\lambda_{\mathrm{cyc}}$ around their default values. Across all settings, the harmonic mean changes by less than 0.5\%, indicating that the final performance is largely insensitive to these hyperparameters.

\paragraph{Gating configurations.}
We compare full gating (length + depth), length-only gating, and depth-only gating. These variants exhibit nearly identical performance, suggesting that the failure is not caused by a specific gating submodule but by the broader optimization difficulty of prompt-level gating in this setting.

\subsubsection{Gradient Balancing Attempts}
\label{sec:gradient_balancing}

To test whether the observed failure can be rescued by standard optimization tricks, we explored several gradient-balancing strategies:

\begin{itemize}
    \item \textbf{Learning-rate scaling:} assigning a 50$\times$ larger learning rate to gate parameters;
    \item \textbf{Gradient clipping:} clipping gradients with max norm 1.0;
    \item \textbf{Alternative initialization:} zero, uniform, and biased initializations for gate logits;
    \item \textbf{Adaptive optimizers:} replacing SGD with AdamW under multiple learning-rate and weight-decay settings;
    \item \textbf{Temperature annealing:} applying a temperature-controlled sigmoid together with entropy regularization;
    \item \textbf{Phased training:} separating warm-up, gate-only adaptation, and joint optimization stages.
\end{itemize}

None of these strategies restored meaningful gate diversity or improved performance beyond the fixed-prompt baselines. In some cases, such as aggressive initialization changes or adaptive optimizers, training became unstable and degraded sharply.

\subsubsection{Attempts to Revive Adaptive Gating}
\label{sec:revive_gating}

We also tested two more direct repair strategies motivated by the diagnosed failure modes.

\paragraph{Gradient equilibrium mechanism.}
We introduced a scaling factor intended to offset sigmoid-induced attenuation:
\[
\tilde{g}_i = \alpha_i \frac{\partial \mathcal{L}}{\partial g_i},
\qquad
\alpha_i=\frac{1}{\sigma(g_i)(1-\sigma(g_i))+\epsilon},
\]
with $\epsilon=10^{-8}$ and a maximum gradient scale of 10.0.

\paragraph{Entropy regularization.}
We added an entropy-based loss encouraging gate outputs away from saturated states:
\[
\mathcal{L}_{\mathrm{ent}}
=
\lambda\sum_i \big[
\sigma(g_i)\log\sigma(g_i)
+
(1-\sigma(g_i))\log(1-\sigma(g_i))
\big].
\]

\paragraph{Observation.}
Although the gradient equilibrium mechanism numerically reduced the magnitude gap, it did not restore substantial gate variation during training. Effective prompt lengths and depth activation probabilities remained near-constant, and performance gains did not materialize. Entropy regularization likewise failed to produce useful gate diversity.

These results support the interpretation that the collapse is structural rather than a consequence of a single missing optimization heuristic.

\subsection{Additional Analysis and Clarifications}

\subsubsection{EuroSAT Controlled Variants}
\label{sec:dataset_behavior_supp}

To understand why AdaptiveBiMaPLe performs relatively better on EuroSAT than on ImageNet or Caltech101, we evaluated controlled variants designed to separate adaptive behavior from parameter-count and regularization effects:

\begin{itemize}
    \item \textbf{ParamMatched:} remove adaptive gates while adding comparable trainable parameters;
    \item \textbf{AlwaysFrozen:} keep gates but freeze them from the beginning;
    \item \textbf{ExplicitReg:} replace adaptive behavior with explicit regularization such as dropout and weight decay.
\end{itemize}

The main paper reports that ParamMatched and ExplicitReg achieve comparable or higher performance than the original adaptive model. These results indicate that the EuroSAT improvement does not uniquely depend on dynamic gating. Instead, it is more consistent with parameter buffering and regularization effects in small-data regimes. While domain-specific factors may still contribute, adaptive gating itself is not necessary to obtain the observed gain.

\subsubsection{On Gradient Ratio and Seed Variance}

Reviewer comments asked what magnitude ratio should be considered ``healthy'' for adaptive behavior. We do not claim a single universal threshold. Instead, we consider gradients healthy when gate parameters receive signals comparable in scale to other trainable parameters and continue to exhibit sustained variation over training. In our experiments, persistent 2--3 order magnitude gaps consistently coincide with early gate saturation, near-constant activation statistics, and no meaningful functional benefit.

The higher cancellation rate observed for some seeds should therefore not be interpreted as evidence of robust adaptation. Rather, it reflects the instability of an already weak signal: when gradients are extremely small, minibatch-level fluctuations can dominate the apparent direction of gate updates without changing the eventual collapsed behavior.

\subsubsection{Relation to MoE-Style Gating}

Our conclusions should not be conflated with results on gating in mixture-of-experts (MoE) systems. In MoE architectures, gating routes among high-capacity expert networks whose parameters are fully trainable and receive strong gradient signals through deep activation paths. In contrast, our setting studies prompt-level gating over low-dimensional additive prompt tokens on frozen backbones. This structural difference leads to much weaker gradient flow to gates in prompt-only tuning and explains why prompt gating can fail even when gating is effective in other architectural contexts.

Accordingly, our paper does not argue that gating mechanisms universally fail. The diagnostic conclusion is specific to prompt-level sigmoidal gating under frozen few-shot prompt learning.

\end{document}